\RequirePackage[T1]{fontenc}
\documentclass[conference]{IEEEtran}
\IEEEoverridecommandlockouts
\usepackage{cite}
\usepackage{graphicx}
\usepackage{amsmath,amssymb}
\usepackage{url}
\usepackage{booktabs}
\usepackage{multirow}
\ifdefined\pdfminorversion
\else
    
\fi

\title{MCUFlowNet: Hardware-Aware Neural Architecture Search for Optical Flow on Microcontrollers}
\author{\IEEEauthorblockN{Enmin Lin, Jona Beysens}
\IEEEauthorblockA{Department of Electrical Engineering (ESAT)\\
KU Leuven, Leuven, Belgium\\
\{enmin.lin, jona.beysens\}@kuleuven.be}\thanks{The resources and services used in this work were provided by the VSC (Flemish Supercomputer Center), funded by the Research Foundation -- Flanders (FWO) and the Flemish Government.}}

\begin{document}
\raggedbottom
\maketitle
\begin{abstract}
Dense optical flow allows small drones to avoid obstacles. However, few dense optical flow networks are designed for microcontrollers (MCUs) with neural accelerators, whose on-chip SRAM limits the input resolution.
We present PACE-NAS, a macro-to-micro neural architecture search framework that decouples memory from latency: a macro design fixes the backbone and its peak SRAM, and a micro design searches operators and depths to trade accuracy against compiler-estimated latency.
We apply it to EdgeFlowNet on a Grove Vision AI V2 board (Ethos-U55), where profiling locates the peak SRAM at the full-resolution output head.
Replacing transposed convolutions with bilinear upsampling and convolutions lowers the peak SRAM by 18\%, and two lightweight context modules lower the endpoint error (EPE) without raising the peak.
All searched subnets share this peak, so the search needs no memory constraint.
We name the Pareto-optimal models MCUFlowNet; the fastest variant, MCUFlowNet-S, runs 1.76$\times$ faster than a retrained EdgeFlowNet at a 1.6\% higher EPE and, in simulated forest flights at 2~m/s, reaches the goal without collision in 51.5\% of trials, against 41.8\%.
PACE-NAS also applies to semantic segmentation, where a searched PIDNet-S variant runs 1.17$\times$ faster at similar accuracy.
Code is available at \url{https://gitlab.kuleuven.be/emedia-research-lab/esat-stadius/mcuflownet/mcuflownet}.
\end{abstract}

\section{Introduction}
\label{sec:introduction}

Small aerial robots rely on timely dense optical flow, the per-pixel displacement between consecutive frames, for collision avoidance~\cite{Bouwmeester2023NanoFlowNet}, and on optical flow for autonomous landing and self-motion estimation~\cite{ParedesValles2024Neuromorphic}.
Perception delay shortens the time available to avoid obstacles~\cite{Falanga2019Latency}, which motivates onboard processing~\cite{Palossi2022Frontnet}, and strict weight and power budgets favor microcontrollers (MCUs)~\cite{vanDijk2024RouteFollowing}.

EdgeFlowNet, a compact multi-scale network of standard convolutions, brings dense optical flow to the Coral Edge TPU~\cite{Raju2025EdgeFlowNet}, but on MCUs its full-resolution output head sets the peak SRAM and thus bounds the input resolution.
Hardware-aware neural architecture search (NAS) adapts networks to a target device~\cite{Tan2019MnasNet}; on MCUs, it treats peak SRAM as a constraint and can be co-designed with the memory scheduling of the inference engine~\cite{Lin2020MCUNet,Lin2021MCUNetV2}.
For optical flow, however, NAS has typically targeted GPU-scale models~\cite{Lin2024FlowNAS}, and few dense flow networks are designed for MCUs with neural accelerators.

\begin{figure}[t]
    \centering
    \includegraphics[width=\columnwidth]{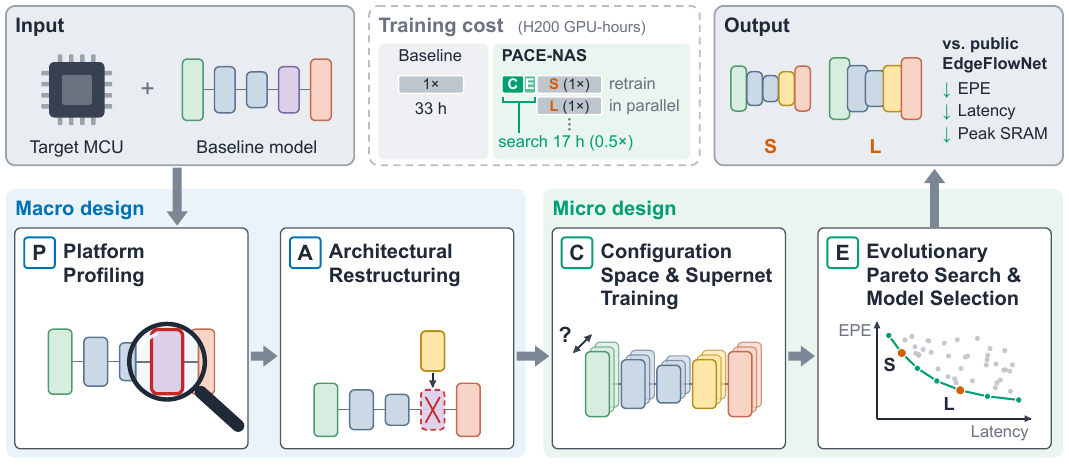}
    \caption{PACE-NAS overview. Training cost is in estimated H200 GPU-hours, relative to one full training run of EdgeFlowNet (1$\times$); S and L mark the two retrained subnets.}
    \label{fig:pace-overview}
\end{figure}

This paper proposes PACE-NAS, a macro-to-micro framework for hardware-aware NAS that adapts existing dense flow networks to MCUs by decoupling memory from latency (Fig.~\ref{fig:pace-overview}).
Its macro design profiles the target platform, restructures the layers that set the peak SRAM, and adds two lightweight context modules for image-wide context where SRAM is spare; these steps fix the backbone and its peak SRAM.
Its micro design trains a weight-sharing supernet over operators and depths that leave this peak unchanged, then runs an evolutionary Pareto search over endpoint error (EPE) and the latency estimated by Arm's Vela compiler~\cite{Arm2025Vela450}.
To the best of our knowledge, this is the first work to use hardware-aware NAS to design dense optical flow networks for MCUs.

On Grove Vision AI V2, a board with an Arm Cortex-M55 MCU and an Ethos-U55 accelerator, we apply PACE-NAS to EdgeFlowNet and retrain two Pareto-optimal subnets, which we call MCUFlowNet-S and MCUFlowNet-L.
Both our models need less peak SRAM than a retrained EdgeFlowNet; MCUFlowNet-S is 1.76$\times$ faster at a 1.6\% higher EPE, while MCUFlowNet-L matches EdgeFlowNet's latency at a similar EPE.
Mainly because of its lower latency, MCUFlowNet-S also raises the success rate in simulated obstacle avoidance from 41.8\% to 51.5\%.
On semantic segmentation, PACE-NAS finds a faster PIDNet-S at similar accuracy.

\begin{figure*}[t]
    \centering
    \includegraphics[width=\textwidth]{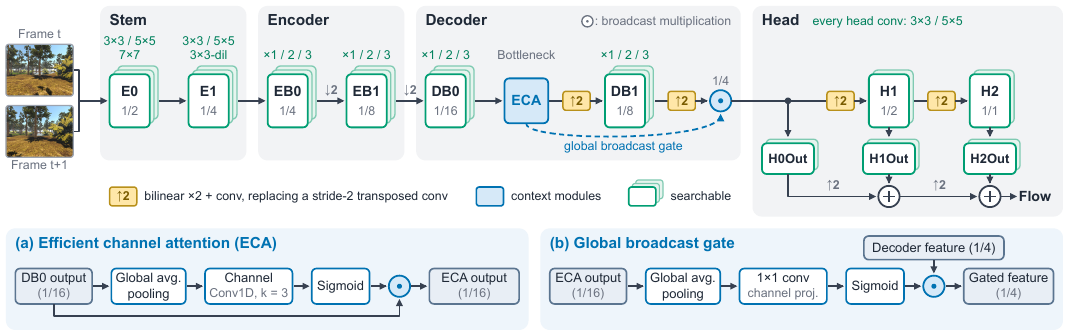}
    \caption{MCUFlowNet backbone and configuration space. $\times1$, $\times2$, or $\times3$ gives the number of residual blocks, and $3\times3$-dil combines a stride-2 $3\times3$ convolution with a dilation-2 $3\times3$ convolution. Insets (a) and (b) detail ECA and the global broadcast gate.}
    \label{fig:backbone-space}
\end{figure*}

\section{Related Work}
\label{sec:related-work}
Accurate optical flow networks typically run on GPUs: RAFT refines flow recurrently from all-pairs correlations~\cite{Teed2020RAFT}, and NeuFlow~v2 reaches real-time operation on embedded GPUs with coarse global matching and lightweight recurrent updates~\cite{Zhang2025NeuFlowV2}.
Lighter pyramid networks estimate flow from coarse to fine: SPyNet warps one frame toward the other at each level and predicts a residual flow~\cite{Ranjan2017SPyNet}, and FastFlowNet adds a compact cost volume, which stores matching costs between features of the two frames~\cite{Kong2021FastFlowNet}.
On MCUs, classical block matching has been accelerated with multicore parallelism~\cite{Kuehne2022ParallelFlow}, and NanoFlowNet runs on GAP8 with depthwise separable convolutions~\cite{Bouwmeester2023NanoFlowNet}.
EdgeFlowNet instead uses an encoder--decoder of standard convolutions, without cost volumes or warping, and accumulates flow increments predicted at three scales~\cite{Raju2025EdgeFlowNet}; all its operators run on the Ethos-U55, so we build on EdgeFlowNet.

NAS has not yet been applied to dense optical flow on MCUs: FlowNAS searches optical flow encoders for GPU-scale models under a parameter budget~\cite{Lin2024FlowNAS}, and multi-objective zero-shot NAS returns a Pareto set for image classification~\cite{Narduzzi2025SteerableNAS}.
NAS on MCUs has targeted classification and detection, where peak SRAM also depends on how the inference engine schedules layers in memory. MCUNet first fits its search space to the memory budget, then searches within it and co-designs the inference engine~\cite{Lin2020MCUNet}; MCUNetV2 jointly searches networks and patch schedules to lower peak SRAM, at the cost of recomputing overlapping features~\cite{Lin2021MCUNetV2}.
Like MCUNet, PACE-NAS designs before it searches, but it first lowers peak SRAM by restructuring the backbone and then searches only choices that leave this peak unchanged, so the search needs no memory constraint.

\section{PACE-NAS: A Macro-to-Micro Design Framework}
\label{sec:method}

Fig.~\ref{fig:pace-overview} outlines the four steps of PACE-NAS: platform profiling (P) and architectural restructuring (A) form the macro design, which fixes the backbone and its peak SRAM; supernet training over a configuration space (C) and evolutionary Pareto search (E) form the micro design.
Fig.~\ref{fig:backbone-space} shows the resulting backbone for optical flow; the search changes only its 11 searchable boxes.

\subsection{Platform Profiling and Bottleneck Diagnostics (P)}
On Grove Vision AI V2, firmware and frame buffers leave 1434~KiB of SRAM available for the tensor arena holding intermediate tensors. Fig.~\ref{fig:resource-profile} presents per-stage Vela profiles of EdgeFlowNet in 8-bit integer (INT8) precision at $160\times208$, before and after restructuring: before, the peak SRAM reaches 1430~KiB at the full-resolution output head (top), only 4~KiB below the available budget. This peak arises because the head must hold its upsampled input and its output simultaneously. Together, this head and the transposed convolution feeding it also account for about 40\% of the estimated cycles, so we target the transposed convolutions for restructuring.

\subsection{Architectural Restructuring (A)}
\begin{figure}[t]
    \centering
    \includegraphics[width=0.9\columnwidth,trim=0 12 0 5,clip]{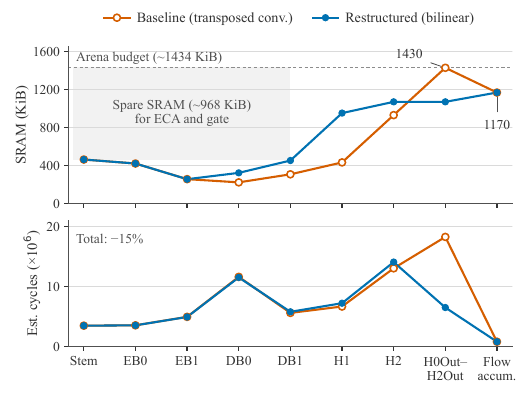}
    \caption{Per-stage Vela profiles of the seed-42 INT8 baseline and restructured models (Table~\ref{tab:ablation}) at $160\times208$: peak SRAM (top) and estimated cycles including memory access (bottom). Restructuring lowers the peak by 18\% and moves it from the output head to final flow upsampling, leaving spare SRAM in the early stages. Stage names follow Fig.~\ref{fig:backbone-space}.}
    \label{fig:resource-profile}
\end{figure}

Vela~4.5.0 can cascade compatible layers (e.g., convolutions), computing them stripe by stripe so that the feature maps between them are never stored whole, but it does not cascade transposed convolutions~\cite{Arm2025Vela450}.
We therefore replace all 11 transposed convolutions with standard convolutions of the same kernel size and insert a bilinear $\times2$ resize before each of the four former stride-2 layers ($\uparrow$2 in Fig.~\ref{fig:backbone-space}), but keep EdgeFlowNet's coarse-to-fine accumulation of flow increments.
The peak then moves to final flow upsampling (Flow accum.\ in Fig.~\ref{fig:resource-profile}), and the restructured backbone fits a $172\times224$ input, at which EdgeFlowNet exceeds the tensor arena.

Because each convolution sees only a local window~\cite{Hu2018SENet}, we use the spare SRAM in the early stages (Fig.~\ref{fig:resource-profile}) to add image-wide context before flow prediction.
Efficient channel attention (ECA)~\cite{Wang2020ECANet} reweights the channels of the lowest-resolution feature, at $1/16$ of the input resolution (DB0 in Fig.~\ref{fig:backbone-space}), where features carry broad spatial context and the tensor is small.
A global broadcast gate adapted from squeeze-and-excitation~\cite{Hu2018SENet} then uses this context to modulate the $1/4$ decoder feature, which feeds all three flow heads (Fig.~\ref{fig:backbone-space}).

\subsection{Configuration Space and Supernet Training (C)}
The configuration space of PACE-NAS (Fig.~\ref{fig:backbone-space}) covers convolutional operators and residual-block depths and fixes connections, channel counts, upsamplers, ECA, and the gate.
It contains no depthwise convolutions: on the Ethos-U55, $3\times3$ depthwise convolutions reach at most about 10\% of peak throughput, whereas standard $3\times3$ convolutions exceed 90\% once channel counts reach 32~\cite{Zhao2025H4H}.
Unlike channel counts, its choices leave tensor sizes unchanged and change only layers that need less SRAM than final flow upsampling, where the peak lies; no subnet can raise the peak, so the search needs no memory constraint (Section~\ref{sec:analysis}).

E0 and E1 each choose among three convolutional operators, which set the spatial context passed downstream; EB0, EB1, DB0, and DB1 each hold one to three residual blocks, which set capacity across scales; and each of the five costly high-resolution head convolutions uses a $3\times3$ or $5\times5$ kernel for local context (Fig.~\ref{fig:backbone-space}).
Together, these choices give $3^6\times2^5=23{,}328$ subnets.

We train a weight-sharing supernet on FlyingChairs2 (FC2)~\cite{Ilg2018Occlusions} by sampling single paths, as in Single Path One-Shot (SPOS)~\cite{Guo2020SPOS}.
Because subnets are scored with these shared weights instead of being retrained, we use FairNAS-inspired balanced sampling~\cite{Chu2021FairNAS} to give competing operators comparable training exposure.
Because the paths share weights, each sampled path pulls the shared layers toward its own features during supernet training~\cite{Lin2024FlowNAS}. We therefore add FlowNAS's feature alignment distillation to the multi-scale flow and uncertainty losses~\cite{Raju2025EdgeFlowNet}: at corresponding layers, every path learns to match a frozen EdgeFlowNet teacher, so all paths move toward the same features.

\subsection{Evolutionary Pareto Search and Model Selection (E)}
As in SPOS, we score each subnet by its FC2 validation EPE with inherited supernet weights (supernet EPE) and by the latency that Vela estimates from its exported INT8 graph.
NSGA-II~\cite{Deb2002NSGAII} then searches for the EPE--latency Pareto front.
From this front, we retrain two subnets from scratch and deploy them in INT8 (Section~\ref{sec:evaluation}): MCUFlowNet-S, the fastest, and MCUFlowNet-L, the closest to EdgeFlowNet in Vela-estimated latency at $172\times224$, for an equal-latency comparison.
\section{Experimental Evaluation}
\label{sec:evaluation}

\subsection{Experimental Setup}
The supernet is trained on FC2 at $172\times224$ for 200 epochs, after which NSGA-II evaluates 800 of the 23{,}328 subnets; together, these take an estimated 17 H200 GPU-hours, about half of one full EdgeFlowNet training run (Fig.~\ref{fig:pace-overview}).

For Table~\ref{tab:deployment}, we train MCUFlowNet-S, MCUFlowNet-L, and EdgeFlowNet from scratch at $160\times208$, the largest input at which EdgeFlowNet fits the tensor arena, with the same recipe: 400 epochs on FC2, then 10k steps on random FC2 crops resized to the input and 80k steps on a 3:1 mix of FC2 crops and FlyingThings3D (FT3D) frames~\cite{Mayer2016FlyingThings}.
For every model trained on FT3D, we keep the checkpoint with the lowest FT3D validation EPE.
Test EPE is computed on all pairs of the Final pass of the MPI-Sintel training set~\cite{Butler2012Sintel} at its central $416\times1024$ crop, without clipping large ground-truth displacements; deployed models see resized inputs, and their outputs are rescaled to the crop.
The deployed models are quantized post-training to INT8 with 64 FC2 calibration pairs and compiled with Vela 4.5.0 for an Ethos-U55 with 64 multiply--accumulate (MAC) units at 400~MHz.
Latencies in Tables~\ref{tab:deployment} and~\ref{tab:segmentation} are measured on the board with TensorFlow Lite Micro~\cite{David2021TFLM}, excluding image capture and pre- and post-processing.

\subsection{Macro Design Sets Memory, Search Allocates Latency}
\label{sec:analysis}
Table~\ref{tab:ablation} ablates the macro-design components on EdgeFlowNet: the restructured decoder, ECA, and the gate.
At $172\times224$, restructuring reduces the Vela-estimated peak SRAM by 18\%, to fit within the tensor arena, and latency by 15\%, while the mean EPE changes by less than one standard deviation.
ECA and the gate each lower EPE for every seed.
Together they add 0.9\% Vela-estimated latency without raising peak SRAM, and MCUFlowNet keeps both.

\begin{table}[t]
\caption{Decoder ablation on EdgeFlowNet trained on FC2 and FT3D at $352\times480$: FP32 Sintel Final EPE (mean $\pm$ SD, three seeds) and Vela estimates at $172\times224$.}
\label{tab:ablation}
\centering
\scriptsize
\setlength{\tabcolsep}{3.5pt}
\begin{tabular}{@{}lrrr@{}}
\toprule
Decoder variant & EPE (px)$\downarrow$ & SRAM (KiB)$\downarrow$ & Latency (ms)$\downarrow$ \\
\midrule
Transposed convolution & 6.96\,$\pm$\,0.03 & 1694\rlap{$^{*}$} & 195.2 \\
Bilinear resize + convolution & 6.98\,$\pm$\,0.06 & \textbf{1386} & \textbf{165.0} \\
\quad + ECA & \textbf{6.76}\,$\pm$\,\textbf{0.05} & \textbf{1386} & 165.3 \\
\quad + gate & 6.83\,$\pm$\,0.06 & \textbf{1386} & 166.4 \\
\quad + ECA + gate & 6.78\,$\pm$\,0.04 & \textbf{1386} & 166.5 \\
\bottomrule
\multicolumn{4}{@{}l@{}}{$^{*}$Exceeds the tensor arena.}
\end{tabular}
\end{table}

All 800 evaluated subnets share the same peak SRAM but span $2.19\times$ in Vela-estimated latency, so the macro design alone sets the peak, and the search only decides where to spend latency.
On the Ethos-U55, latency is better spent on depth at low resolution than on large high-resolution kernels such as EdgeFlowNet's.
An extra residual block at $1/8$ or $1/16$ resolution adds 9--24~ms and lowers the supernet EPE by 0.08--0.21~px, whereas a $5\times5$ instead of a $3\times3$ kernel at H1 or H2 adds about 12~ms but lowers it by only 0.001--0.002~px.
Accordingly, 50 of the 51 Pareto-optimal subnets use $3\times3$ kernels at H1 and H2, and MCUFlowNet-L spends 95\% of its extra latency over MCUFlowNet-S on these low-resolution residual blocks.
Vela's latency estimates are reliable for the search: on the board, all deployed models run within 2.1\% of their estimates.

\subsection{MCUFlowNet Runs Faster or as Fast at Similar EPE}
\label{sec:benchmark}
Table~\ref{tab:benchmark} reports MCUFlowNet and published FP32 optical flow models with public weights.
Models with cost volumes or warping reach the lowest EPEs, but in our tests none of them, not even FastFlowNet and SPyNet with the fewest parameters, could be deployed on the MCU.
Compared with EdgeFlowNet, MCUFlowNet-S lowers the EPE by 17\% at a similar MAC count, and MCUFlowNet-L by 29\% with 1.5$\times$ the MACs.

\begin{table}[t]
\caption{FP32 optical flow models on the Final pass of the MPI-Sintel training set at $416\times1024$. Iteration counts are in parentheses.}
\label{tab:benchmark}
\centering
\scriptsize
\setlength{\tabcolsep}{5pt}
\begin{tabular*}{\columnwidth}{@{\extracolsep{\fill}}p{1.8cm}lrrr@{}}
\toprule
Type & Model & Params (M) & MACs (G) & EPE (px)$\downarrow$ \\
\midrule
\multirow{4}{1.8cm}{\raggedright\emph{Cost volume or warping}} & RAFT (32)~\cite{Teed2020RAFT} & 5.26 & 762.4 & 2.77 \\
 & NeuFlow v2 (9)~\cite{Zhang2025NeuFlowV2} & 9.03 & 109.1 & \textbf{2.59} \\
 & FastFlowNet~\cite{Kong2021FastFlowNet} & 1.37 & 12.7\rlap{$^{*}$} & 4.13 \\
 & SPyNet~\cite{Ranjan2017SPyNet} & 1.44 & 136.2 & 5.59 \\
\midrule
\multirow{3}{1.8cm}{\raggedright\emph{\mbox{Encoder--decoder,} no cost volume or warping}} & EdgeFlowNet~\cite{Raju2025EdgeFlowNet} & 2.74 & 18.6 & 8.20 \\
 & MCUFlowNet-S$^{\dagger}$ & 2.65 & 17.8 & 6.80 \\
 & MCUFlowNet-L$^{\dagger}$ & 5.90 & 28.4 & \textbf{5.79} \\
\bottomrule
\end{tabular*}
\par\vspace{1pt}
\parbox{\columnwidth}{$^{*}$Input resized to $448\times1024$, as FastFlowNet needs sides divisible by 64. $^{\dagger}$Trained with EdgeFlowNet's two-stage scheme (FC2 at $352\times480$, then FT3D at $480\times640$), adding cosine learning-rate decay, gradient clipping, and FT3D augmentation.}
\end{table}

Table~\ref{tab:deployment} reports the deployed INT8 models on the board. Our recipe reduces the EPE of EdgeFlowNet by 20\% relative to its public weights, so we compare against this retrained baseline.
At the same input, MCUFlowNet-S runs 1.76$\times$ faster at a 1.6\% higher EPE, and MCUFlowNet-L matches the baseline's latency at a similar EPE.
Both MCUFlowNet models need the same peak SRAM, 18\% less than EdgeFlowNet, so choosing between them trades only speed against accuracy.

\begin{table}[b]
\caption{INT8 models on Grove Vision AI V2 at $160\times208$. SRAM: Vela-estimated peak. SR and Coll.: success rate and collisions per trial over the same 194 simulated trials for every model, at 2~\textnormal{m/s}.}
\label{tab:deployment}
\centering
\scriptsize
\setlength{\tabcolsep}{1pt}
\begin{tabular*}{\columnwidth}{@{\extracolsep{\fill}}lrrrrr@{}}
\toprule
Model & EPE (px)$\downarrow$ & SRAM (KiB)$\downarrow$ & Latency (ms)$\downarrow$ & SR (\%)$\uparrow$ & Coll.$\downarrow$ \\
\midrule
EdgeFlowNet (public) & 8.03 & 1430 & 164.7 & 45.4 & 0.88 \\
EdgeFlowNet (retrained) & 6.40 & 1430 & 164.3 & 41.8 & 0.92 \\
MCUFlowNet-S & 6.50 & \textbf{1170} & \textbf{93.2} & \textbf{51.5} & \textbf{0.77} \\
MCUFlowNet-L & \textbf{6.36} & \textbf{1170} & 163.6 & 45.4 & 0.87 \\
\midrule
MCUFlowNet-S, slowed$^{*}$ & -- & -- & 164.3 & 43.8 & 0.91 \\
\bottomrule
\end{tabular*}
\par\vspace{1pt}
\parbox{\columnwidth}{$^{*}$Steering interval set to the latency of the retrained EdgeFlowNet.}
\end{table}

\subsection{Lower Latency Improves Obstacle Avoidance}
\label{sec:closed-loop}
We fly a simulated quadrotor through the AvoidBench forest~\cite{Yu2023AvoidBench} at 2~m/s; a trial succeeds if the quadrotor reaches the goal without collision.
As in EdgeFlowNet~\cite{Raju2025EdgeFlowNet}, the controller steers toward free space near the goal direction.
All flow models share this controller, and each steering update waits for the model's Vela-estimated latency plus link and frame delays.
MCUFlowNet-S succeeds more often than the retrained EdgeFlowNet (51.5\% against 41.8\%; Table~\ref{tab:deployment}).
When slowed to the steering interval of the retrained EdgeFlowNet, it succeeds in only 43.8\% of trials and collides about as often as the baseline (0.91 against 0.92 per trial), so most of this gain comes from its lower latency, which lets the quadrotor react to obstacles earlier.
At EdgeFlowNet's latency, MCUFlowNet-L and the two EdgeFlowNet models succeed in 41.8\% to 45.4\% of trials whatever their EPE (6.36 to 8.03~px), so in this task latency matters more than EPE.

\subsection{Generalization to Segmentation}
\label{sec:segmentation}
Beyond optical flow, we apply PACE-NAS to PIDNet-S~\cite{Xu2023PIDNet} on Cityscapes~\cite{Cordts2016Cityscapes} at a $160\times320$ input, at which PIDNet-S already peaks at 1350~KiB, 84~KiB below the tensor arena, so the macro design skips restructuring and adds only ECA and the gate. The configuration space covers the number of residual blocks in each of the seven stages and, in six of them, the second-convolution kernel size ($3\times3$ or $5\times5$) of every block after the first. NSGA-II evaluates 960 of the 1.41 million subnets by INT8 mean intersection over union (mIoU) on the validation images and by Vela-estimated latency. All models start from the same ImageNet weights and use the official recipe with training views cropped from the full-resolution image, as for optical flow; we train three seeds per model and quantize them post-training with 256 training images. From the Pareto front, we retrain an accurate subnet and the subnets faster than PIDNet-S; Table~\ref{tab:segmentation} reports three of them: a fast, a balanced, and an accurate subnet.

\begin{table}[!h]
\caption{INT8 segmentation on Grove Vision AI V2 at $160\times320$ (500 Cityscapes validation images; mean over three seeds, $\pm$ SD for mIoU); all models peak at 1350~\textnormal{KiB}.}
\label{tab:segmentation}
\centering
\scriptsize
\begin{tabular}{@{}lrr@{}}
\toprule
Model & mIoU (\%)$\uparrow$ & Latency (ms)$\downarrow$ \\
\midrule
PIDNet-S (retrained) & 55.23\,$\pm$\,0.55 & 188.6 \\
PIDNet-S + ECA + gate & 55.44\,$\pm$\,0.73 & 191.4 \\
\midrule
Searched (fast) & 54.22\,$\pm$\,0.68 & \textbf{138.9} \\
Searched (balanced) & 55.27\,$\pm$\,0.60 & 161.0 \\
Searched (accurate) & \textbf{55.81}\,$\pm$\,\textbf{0.54} & 200.4 \\
\bottomrule
\end{tabular}
\end{table}
Adding ECA and the gate raises the mean mIoU of the retrained PIDNet-S slightly, by 0.2 points (Table~\ref{tab:segmentation}).
Relative to the retrained PIDNet-S, the fast subnet trades 1.0 point of mIoU for a 1.36$\times$ speedup, the balanced subnet matches its mIoU (+0.04 points on average) with a 1.17$\times$ speedup, and the accurate subnet gains only 0.6 points, about one standard deviation, at 6\% higher latency.
The search can thus trade off accuracy against latency on segmentation as well.

\section{Conclusion}
PACE-NAS treats peak SRAM as a design problem and latency as a search problem.
Profiling (P) locates the layer that sets the peak SRAM, restructuring (A) lowers this peak once, for all subnets, and a configuration space (C) that leaves tensor sizes unchanged keeps it there, so the Pareto search (E) handles only accuracy and latency.
PACE-NAS turns EdgeFlowNet, an existing network designed for the Edge TPU, into MCUFlowNet for the Ethos-U55.
We build two variants, both with 18\% less peak SRAM than EdgeFlowNet: against a retrained EdgeFlowNet, MCUFlowNet-S runs 1.76$\times$ faster at a 1.6\% higher EPE and raises the simulated success rate at 2~m/s from 41.8\% to 51.5\%, while MCUFlowNet-L matches its latency at a similar EPE.
The search also speeds up \mbox{PIDNet-S} for semantic segmentation by 1.17$\times$ at similar mIoU.

\bibliographystyle{IEEEtran}
\bibliography{references}
\end{document}